\documentclass[preprint,12pt]{elsarticle}
\usepackage{amsmath}
\usepackage{booktabs}
\usepackage{float}
\usepackage{graphicx} 
\usepackage{rotating}
\usepackage{booktabs}
\usepackage[ruled,vlined,linesnumbered]{algorithm2e}
\SetKwInput{KwIn}{Input}
\SetKwInput{KwOut}{Output}
\usepackage{amssymb}

\usepackage{amsthm}

\usepackage{lineno}
\usepackage{multirow}
\journal{Information Fusion}
\begin{document}

\begin{frontmatter}



\title{ImageDDI: Image-enhanced Molecular Motif Sequence Representation for
Drug-Drug Interaction Prediction}



\author[1]{Yuqin He \fnmark[+]} 

\author[1]{Tengfei Ma\fnmark[+]}

\author[1]{Chaoyi Li}

\author[1]{Pengsen Ma}

\author[1]{Hongxin Xiang}

\author[2]{Jianmin Wang}

\author[1]{Yiping Liu}

\author[1]{Bosheng Song\cormark[*]} 

\author[1]{Xiangxiang Zeng}

\affiliation[1]{College of Computer science and Electronic Engineering, Hunan University, Changsha, Hunan 410082, china}
\affiliation[2]{Department of Integrative Biotechnology, Yonsei University, Incheon 21983, Republic of Korea}
\cortext[*]{Corresponding author: Bosheng Song (boshengsong@hnu.edu.cn)} 
\fntext[+]{These authors contributed equally to this work.}

\begin{abstract}
To mitigate the potential adverse health effects of simultaneous multi-drug use, including unexpected side effects and interactions, accurately identifying and predicting drug-drug interactions (DDIs) is considered a crucial task in the field of deep learning. Although existing methods have demonstrated promising performance, they suffer from the bottleneck of limited functional motif-based representation learning, as DDIs are fundamentally caused by motif interactions rather than the overall drug structures. In this paper, we propose an Image-enhanced molecular motif sequence representation framework for \textbf{DDI} prediction, called ImageDDI, which represents a pair of drugs from both global and local structures. Specifically, ImageDDI tokenizes molecules into functional motifs. To effectively represent a drug pair, their motifs are combined into a single sequence and embedded using a transformer-based encoder, starting from the local structure representation. By leveraging the associations between drug pairs, ImageDDI further enhances the spatial representation of molecules using global molecular image information (e.g. texture, shadow, color, and planar spatial relationships). To integrate molecular visual information into functional motif sequence, ImageDDI employs Adaptive Feature Fusion, enhancing the generalization of ImageDDI by dynamically adapting the fusion process of feature representations. Experimental results on widely used datasets demonstrate that ImageDDI outperforms state-of-the-art methods. Moreover, extensive experiments show that ImageDDI achieved competitive performance in both 2D and 3D image-enhanced scenarios compared to other models.
\end{abstract}

\begin{graphicalabstract}
\includegraphics[width=\textwidth]{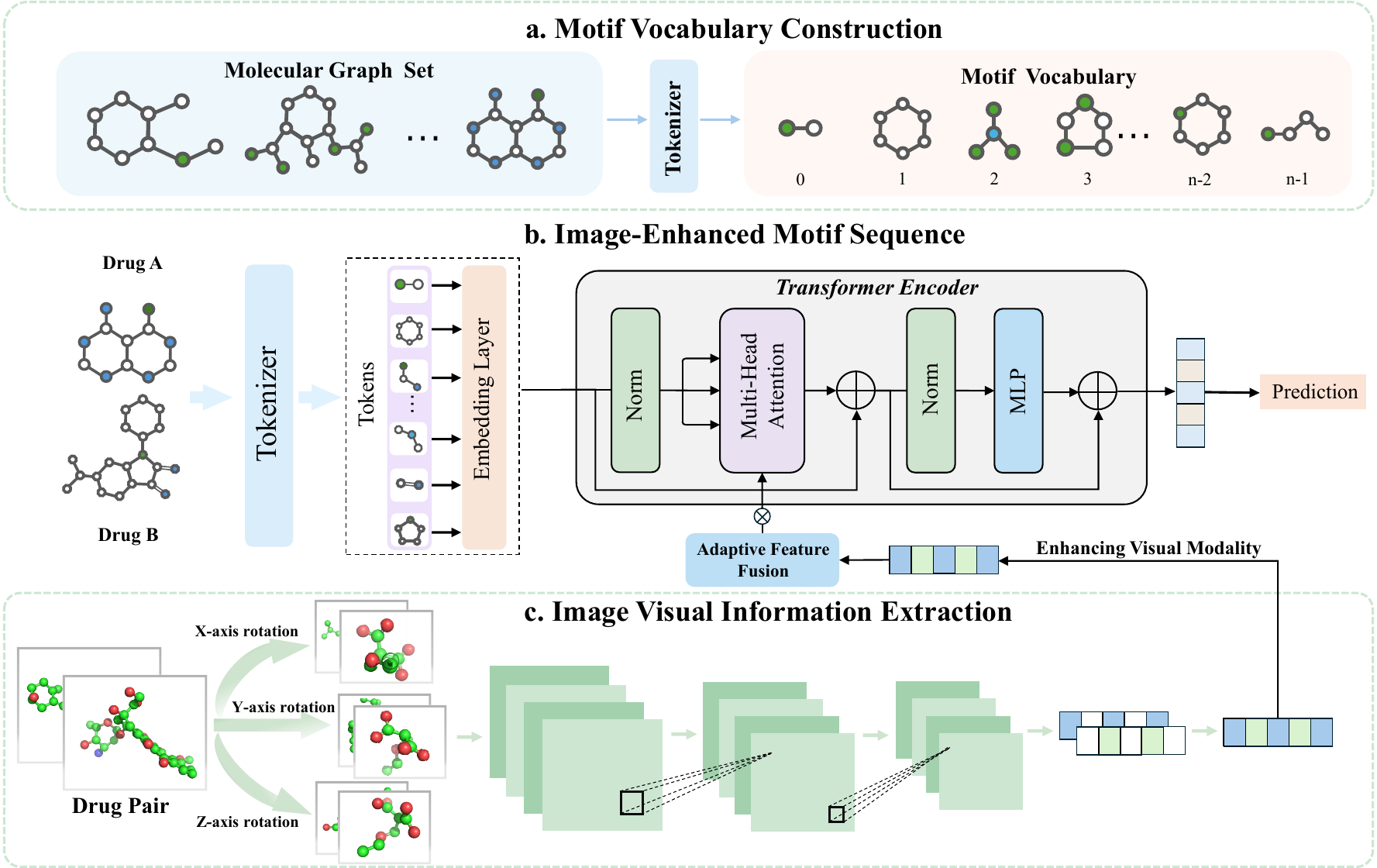}
\end{graphicalabstract}

\begin{highlights}
\item We innovatively proposed a motif-based retrieval method, which constructed a motif vocabulary and characterized the complex relationships between local structures of input drugs.
\item We propose an image-enhanced motif sequence framework that leverages Adaptive Feature Fusion to incorporate global molecular image information, improving motif-based sequence representation. The theoretical effectiveness of the framework is also validated.
\item Extensive experiments on benchmarks demonstrate that ImageDDI outperforms state-of-the-art baseline models. Additionally, it also achieves superior results in inductive scenarios.
\end{highlights}


\end{frontmatter}

\bibliographystyle{unsrt}
\section{Introduction}

The concurrent use of multiple drugs can result in drug-drug interactions (DDIs), which may reduce medication effectiveness and cause severe adverse reactions, endangering patient health~\cite{vilar2014similarity,bansal2014community}. Consequently, early identification of these interactions is vital for ensuring patient safety and improving therapeutic outcomes~\cite{palmer2017combination,giacomini2007good}. Recently, many computational models have been proposed to predict drug-drug interactions~\cite{jin2017multitask,qiu2021comprehensive}. Early methods usually adopt structural similarity profiles and structural information from drug pairs based on deep neural networks (DNNs) to predict DDI types~\cite{ryu2018deep,huang2020caster}. However, these methods adopt DNNs to represent drugs, relying on expert domain knowledge to design efficient features. 

To address this, some advanced methods utilize graph neural networks (GNNs)~\cite{zhang2019graph} to fully extract the structural features of molecules and promote the structural correlations between drugs~\cite{nyamabo2021ssi,nyamabo2022drug,yang2022learning}. Although these methods have achieved promising performance, they focus primarily on the structure of molecules, ignoring the associations between drugs and other biomedical entities~\cite{zhao2023improving,zitnik2018modeling}. Knowledge graphs and biomedical networks effectively integrate and represent multi-source heterogeneous data, providing rich semantic information and clear relational representations \cite{chen2021muffin,yu2021sumgnn,lin2020kgnn}. To further mine the semantic relations and domain knowledge between biomedical entities, a line of work considers knowledge graphs and biomedical networks as the external information to enhance the structural representation of drugs~\cite{wang2022predicting,xiong2023multi,wu2024mkg}. However, for the new drugs in the early stages, the available information between biomedical entities is limited in these methods, which makes it difficult to establish connections with external knowledge graphs and integrate them into interaction graphs. This hinders the model's generalization to new drugs under the inductive scenario. Therefore, some approaches consider fully extracting the structural correlations between molecules~\cite{niu2024srr}. Motifs within molecules determine their pharmacokinetic (how the body processes them) and pharmacodynamic (how they affect the body) properties, ultimately determining all their interactions~\cite{nyamabo2021ssi}.

\begin{figure*}[h]
    \centering
    \includegraphics[width=1\linewidth]{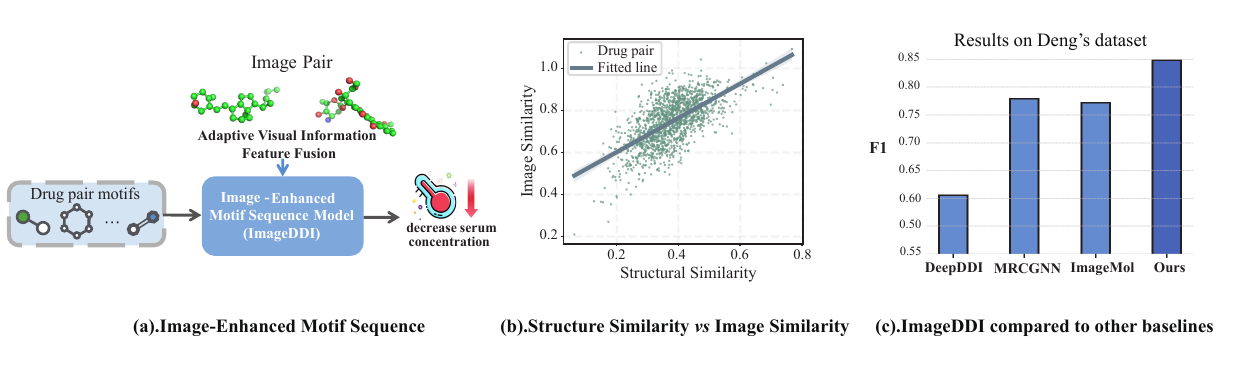}
    \caption{
    (a). ImageDDI representing visual information of molecules enhances the structural representation of drug pairs. (b). The similarity distribution of drug pairs learned from ImageDDI is consistent with the distribution of their molecular structural similarity (Mogan Fingerprint). (c). ImageDDI outperforms other solely graph- or image-based methods.    }
    \label{fig: introduction}
\end{figure*}

Although existing approaches enable substructure-based molecular representation learning, they struggle to effectively model motif interactions between drug pairs~\cite{yang2022learning,tang2023dsil}. To address this, we propose ImageDDI, an image-enhanced molecular motif sequence representation framework for DDI prediction (Figure~\ref{fig: introduction}a). Drug pair motifs are combined into a single sequence and encoded using a Transformer-based encoder~\cite{vaswani2017attention}, starting with local structural representations. To improve motif representation and integrate global molecular structures, we employ \textbf{Adaptive Feature Fusion} to model spatial relationships within local motifs and enhance visual interactions between drugs. As shown in Figure~\ref{fig: introduction}b, the high positive correlation between global molecular images and local motif features (Morgan fingerprint~\cite{morgan1965generation}) demonstrates that ImageDDI effectively captures motif relationships within drug pairs. Figure~\ref{fig: introduction}c further shows that ImageDDI, by incorporating visual information, outperforms graph-based (e.g., MRCGNN) and image-based (e.g., ImageMol) methods in enhancing motif representations.
The main contributions of this work are as follows:
\begin{itemize}
\item We innovatively proposed a motif-based retrieval method, which constructed a motif vocabulary and characterized the complex relationships between local structures of input drugs.

\item We propose an image-enhanced motif sequence framework that leverages Adaptive Feature Fusion to incorporate global molecular image information, improving motif-based sequence representation. The theoretical effectiveness of the framework is also validated.

\item Extensive experiments on benchmarks demonstrate that ImageDDI outperforms state-of-the-art baseline models. Additionally, it also achieves superior results in inductive scenarios.
\end{itemize}

\section{Related Work}

\subsection{Methods based on molecular representation}
Early research often represented drugs as feature vectors, like the DeepDDI~\cite{ryu2018deep} model, which combined structural similarity profiles (SSP) with DNN for DDI prediction. With the rise of GNNs, many studies used GNNs and RDKit to convert SMILES sequences into molecular graphs for DDI prediction. Since DDIs depend on both drug structures and biomedical entities like targets, enzymes, and pathways, many GNN models have integrated external information and used knowledge graphs. Knowledge graph-based models like KGNN \cite{lin2020kgnn} and SumDDI \cite{yu2021sumgnn} have proven effective but rely on supervised data and struggle with new drugs. To improve DDI prediction in inductive scenarios, we propose ImageDDI, which focuses on the intrinsic chemical structure to enhance the generalizability of new drugs.

\subsection{Methods based on molecular substuctures}
Most studies on DDIs rely on partial and incomplete structural information, which highlights the effectiveness of substructure-based approaches. The early SSI-DDI model~\cite{nyamabo2021ssi} employed co-attention with multiple GAT layers~\cite{velivckovic2017graph} to extract node features from diverse receptive fields. SA-DDI~\cite{yang2022learning} further improved this by assigning varying weights to local regions, thereby enhancing the ability to capture substructures. However, these models primarily focus on local substructures within a single molecular graph and fail to treat them as independent entities, leading to a loss of spatial information. The GMPNN model~\cite{nyamabo2022drug} addressed this limitation by treating substructures as independent entities and incorporating a gated mechanism for interactions. Similarly, the DSIL-DDI model~\cite{guo2024dsil} captured fine-grained interactions but overlooked spatial relationships and the global structure of the substructure. Previous methods have generally neglected the structural relationships between drug pairs. To overcome this limitation, we propose ImageDDI, a model that unifies the substructures of drug pairs and incorporates visual-spatial information derived from molecular images to enhance performance.

\cite{zhu2025drug}
\cite{zhu2024drug}
\cite{zhu2023drug}
\cite{zhu2023associative}

\section{Preliminary}
\subsection{Motif Sequence of a Pair of Drugs}
The drug pair \(d_x\) and \(d_y\) are tokenized into \(S_{d_x}\) and \(S_{d_y}\) as the motif sequences of the drug pair. These sequences are merged using an $\oplus$ operation to form a jointed motif sequence $S_{(d_x,d_y)} = S_{d_x} \oplus S_{d_y}$. This motif sequence is then encoded by a Transformer-based model, which begins with local motifs and progressively captures deeper motif interactions, providing a more accurate representation for DDI prediction.

\subsection{ Image-based Molecular Structure Representation}
Recently, molecular imaging has gained attention, with studies using molecular images to extract chemical structures and pixel-level visual data~\cite{zeng2022accurate,xiang2023chemical,10375706}. By adding convolutional and pooling layers, images can represent both spatial structure and global visual information, which is crucial for understanding spatial interactions. Building on molecular dynamics, rotating visual 3D molecular conformers provide an effective way to represent molecular properties. 

\subsection{Problem Formulation}
We define DDI prediction as a multi-class classification task to predict interaction types. To achieve this, we propose an image-enhanced motif sequence model that estimates interaction probabilities with visual guidance. For details, the motif sequence of a drug pair \(d_x\) and \(d_y\) is denoted as \(S(d_x, d_y)\). The model uses \textbf{Adaptive Feature Fusion} to integrate visual information with substructure sequence data for efficient cross-modal representation. The DDI event for each drug pair is predicted as:
\begin{equation}
\hat{y}_{(d_x, d_y)} = \mathcal{F}\left(I_x, I_y \otimes S(d_x, d_y)\right),    
\end{equation}
where \(I_x\) and \(I_y\) are the visual representations of drugs \(d_x\) and \(d_y\)respectively, and the  '\textbf{$\otimes$}' denotes adaptively fuse the fetures.

\section{Method}

\subsection{Overview of the Method}
We propose an image-enhanced motif sequence framework that tokenizes drugs into motifs and integrates global visual information. Specifically, ImageDDI follows the three steps below: (1) Tokenizing drugs into motifs using BRICS and converting drugs into images with RDKit (Section \ref{4.2}); (2) Fusing visual and motif representations in a Transformer (Section \ref{4.3}); (3) Extracting global visual features (Section \ref{4.4}); For a drug set \(\mathcal{D}\), each drug \(d_x \in \mathcal{D}\) is decomposed into motifs to build a motif vocabulary \(\mathcal{V}_{\text{motif}}\). For each pair of drugs \((d_x, d_y)\), their motif sequences \(S_{d_x}\) and \(S_{d_y}\) are concatenated as \(S_{(d_x, d_y)} = S_{d_x} \oplus S_{d_y}\) and processed by a Transformer. The visual features \(I_x\) and \(I_y\) are fused to the Transformer through adaptive feature fusion. The training algorithm is detailed in Appendix A.5.
\begin{figure}[htbp]
    \centering
    \includegraphics[width=0.98\textwidth]{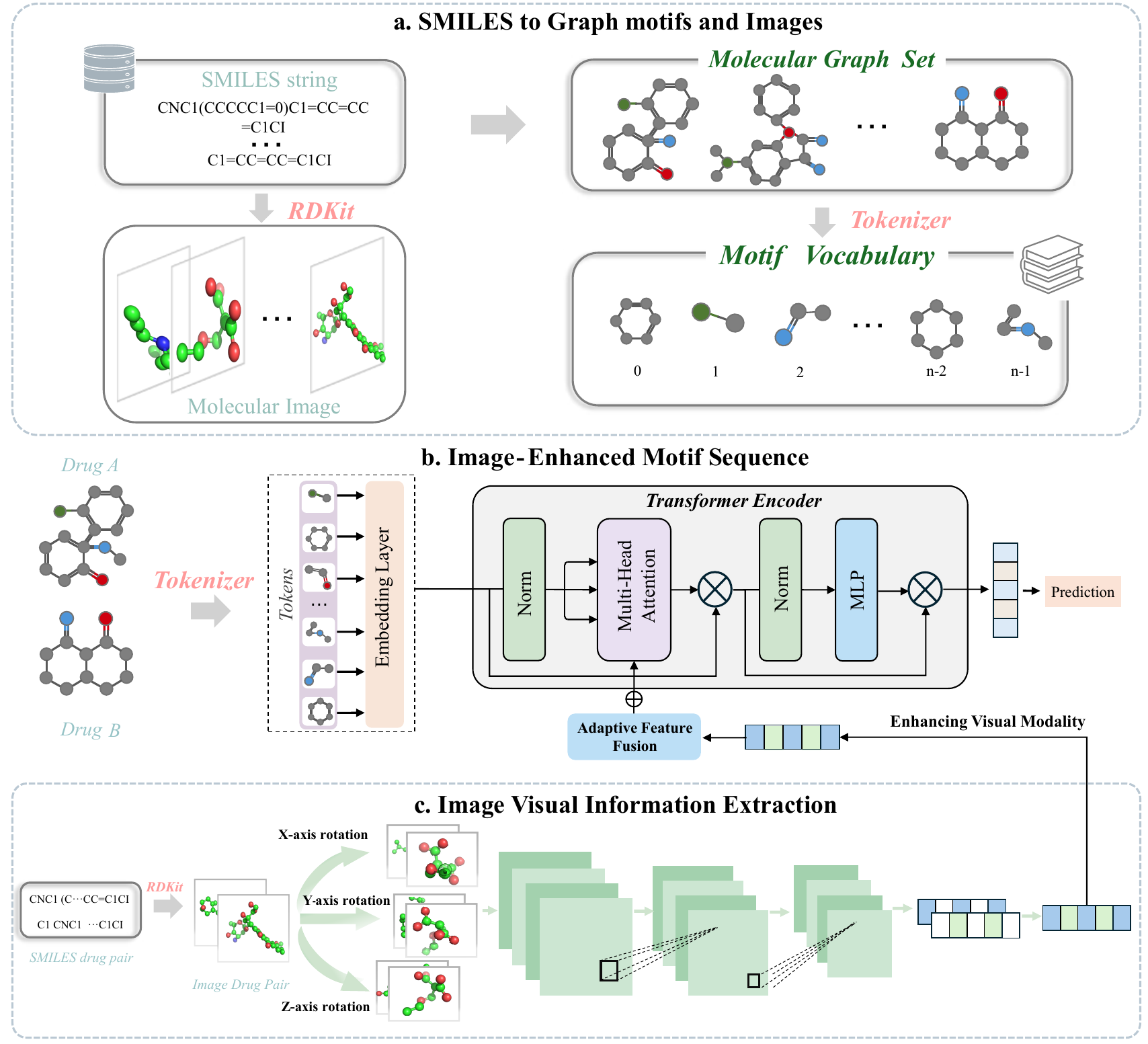}
    \caption{Overview of ImageDDI. (a) Motif Vocabulary Construction: Molecules are tokenized into motifs to build a motif vocabulary.
(b) Image-Enhanced Motif Sequence: A Transformer with adaptive feature fusion integrates image information into motif sequences.
(c) Image Visual Information Extraction: Molecular images are rendered with X-, Y-, and Z-axis rotations to extract visual features.
}
    \label{fig:method}
\end{figure}

\subsection{ SMILES to Graph motifs and Images} \label{4.2}

Motifs function as key local structural that govern molecular behavior in DDIs by enabling specific binding and interaction mechanisms, thereby determining the specificity, affinity, and overall prediction results of DDIs. The ImageDDI framework is designed to align with this concept by split molecular structures to construct motif vocabulary, allows the model to effectively capture the intricate interaction patterns, thereby enhancing the accuracy and reliability of DDI predictions. Molecules are decomposed into functional motifs, which are then used to build a motif vocabulary \(V_\text{motif}\). This method captures essential motif information, providing a strong foundation for downstream tasks and enabling more accurate molecular representations. For the drug set \(\mathcal{D}\), each drug \(d_i \in \mathcal{D}\) is processed with BRICS (for more details in Appendix A.7) to extract its motif set \(S_{d_i}\). BRICS fragments molecules by cleavable bond rules. New motifs \(m_i \in S_{d_i}\) are assigned unique IDs and added to \(V_\text{motif}\), ensuring complete coverage and uniqueness for downstream analysis.

For 2D images, we convert SMILES representations into standardized 2D topological molecular images using RDKit's Chem.Draw module, enforcing consistent rendering parameters (atom/bond visualization schemes and image dimensions) to ensure uniformity in structural feature extraction.

For 3D images, we generate molecular conformers by removing hydrogen atoms and using MMFFOptimizeMolecule() in RDKit with MMFF94 and a maximum of 5000 iterations to generate conformers in a pre-defined coordinate system. If the conformer does not converge, we double the iterations and repeat up to 10 times. If convergence fails after 10 attempts, we use the 2D conformation instead.

\subsection{Image-enhanced Motif Sequence} \label{4.3}
For the input drug pair \((d_x, d_y)\), the motif sequences \(S_{d_x}\) and \(S_{d_y}\) are first extracted and then concatenated into a unified sequence \(S_{(d_x, d_y)} = S_{d_x} \oplus S_{d_y}\). This sequence integrates the local substructure features of both drugs, establishes their overall structural relationships, and provides richer and more reliable input data for downstream tasks. To effectively model motif sequences and integrate visual information for enhanced knowledge sharing between visual and substructure features, we adopt a Transformer architecture, with the \textbf{Adaptive Feature Fusion} module implemented through learnable attention biases. This approach efficiently integrates features from the motif sequence \( S_{(d_x,d_y)} \) and visual information \( I_{xy} \) (see in Section 4.4). To integrate visual modality information into the model through adaptive feature fusion, we introduce bias terms based on visual data. The extended attention formula is:
\begin{equation}
\text{Attention}(Q, K, V) = \text{softmax}\left(\frac{QK^T}{\sqrt{d_k}} + \Phi_{I_{xy}} \cdot I_{xy} \right)V,
\end{equation}
where \(Q = XW_Q\) (query), \(K = XW_K\) (key), and \(V = XW_V\) (value), with \(W_Q, W_K, W_V \in \mathbb {R}^{d \times d_k}\) as learnable weight matrices, and \(d_k\) is the scaling factor to mitigate gradient vanishing. \(\Phi_{I_{xy}}\) is a learnable bias based on the drug pair's visual features \(I_{xy}\), used to adjust the impact of the visual modality in the attention mechanism. It dynamically adjusts the contribution of visual information in feature fusion by extracting features from the visual data, optimizing the interaction between visual and structural features.
 The output of the multi-head attention is further processed by a Feed-Forward Network (FFN), calculated as:

\begin{equation}
 \text{FFN}(x) = \max(0, xW_1 + b_1)W_2 + b_2,   
\end{equation}
where \(W_1, W_2 \in \mathbb{R}^{d \times d}\) are learnable weight matrices, and \(b_1, b_2 \in \mathbb{R}^d\) are learnable biases. The complete computation for each layer is:

\begin{equation}
Z = \text{LayerNorm}(S_{(dx,dy)} + \text{Attention}(Q, K, V)),
\end{equation}
\begin{equation}
Z' = \text{LayerNorm}(Z + \text{FFN}(Z)),
\end{equation}
where \(S_{(dx,dy)}\) is the input motif sequence representation. This approach extends the Transformer by introducing Adaptive Feature Fusion, enhancing its ability to process both sequence and image data. By merging local motif features with visual information, it accurately models molecular interactions, improving DDI prediction.

\subsection{Image Visual Information Extraction} \label{4.4}

While motifs effectively capture local structural information, they lack the entire molecular interaction landscape. We employ image-based techniques to learn global interaction patterns, complementing motif analysis for a more comprehensive understanding of DDIs.

\noindent\textbf{2D Image.} Compared to regular images, molecular images are more sparse, with over 90\% of the area filled with zeros, resulting in "usable" data occupying only a small fraction of the image. Given this limitation, our model does not use "random cropping." We generate molecular images using RDKit, followed by preprocessing and augmentations:  
(1) center cropping to a fixed size;  
(2)  50\% horizontal flipping; 
(3) 20\% grayscale conversion;  
(4) random rotation (0-360°). These methods do not alter the original structure of the molecular images and allow the model to learn invariance to data augmentation.

\noindent\textbf{3D Image.} We use RDKit to optimize molecular conformers, which are rotated counterclockwise (along the x, y, and z axes) at ten angles, generating 10 frames in different orientations. These frames are rendered in stick-ball mode with PyMOL (\(640 \times 480\) RGB images), expanded to \(640 \times 640\), resized to \(224 \times 224\), and stitched into molecular videos \(V = \{I_1, I_2, ..., I_{10}\}\) with resolution \(R^{10 \times 3 \times 224 \times 224}\).

\noindent\textbf{Image Feature Extractor.} We use ResNet18~\cite{he2016deep} as our backbond to encode 2D and 3D images, The 2D encoder extracts from a single 2D molecular image, while the 3D encoder uses view-based mean pooling on multi-view 3D images to obtain features. the image encoder generates visual representations \(I_x\) and \(I_y\) for \(d_x\) and \(d_y\), respectively. These representations are concatenated to form a combined visual representation:
\begin{equation}
I_{xy}= \text{concat}(I_x, I_y),
\end{equation}
This approach captures each molecule's visual features and encodes interaction information through the combined representation, serving as input for DDI prediction.

\subsection{Drug–drug Interaction Prediction}

For each drug pair \( (d_x, d_y) \) and its interaction type \( r \), we now obtain the motif sequence representation \( Z^{\prime} \), which integrates image features processed through a Transformer. This representation is defined as:  
\begin{equation}
h(d_x, d_y, r) = Z^{\prime},
\end{equation}
this representation is processed through a residual layer and a multi-layer perceptron~(MLP), with the final prediction computed as:  
\begin{equation}
\hat{y}_{(d_x,d_y)}^r = \sigma(\phi(\rho(h(d_x, d_y, r)))),
\end{equation}
where \(\sigma\) denotes the softmax function, \(\phi\) represents MLP, and \(\rho\) stands for the residual layer.  the loss function is defined as:  
\begin{equation}
\ell_c = -\frac{1}{N} \sum_{i=1}^{N} \sum_{r=1}^{|R|} y_{(d_x, d_y)}^r \log(\hat{y}_{(d_x, d_y)}^r),
\end{equation}
where \( N \) is the number of samples, \( \hat{y}_{(d_x, d_y)}^r \) is the predicted probability for relation \( r \) and $|R|$ denotes the number of the type of DDI.

\subsection{Justification of ImageDDI Effectiveness}
We use \textbf{Adaptive Feature Fusion} to integrate image information into motif sequences and analyze the information gained from visual features. To formalize the features, we let $F_{(d_x,d_y)}^{IE}$ and $F_{(d_x,d_y)}^M$ denote features from image-enhanced and motif-only sequences. We defined the information gained from visual features as $\mathcal{I}_{gain} = \mathcal{I}(F_{(d_x,d_y)}^{IE} \mid ~S, r; \text{Enc}, \gamma) \texttt{-} \mathcal{I}(F_{(d_x,d_y)}^{M} \mid~S, r; \text{Enc})$, where $\mathcal{I}(\circ \mid *)$ denotes the amount of information produced by $\circ$ under the conditions of given * and $\mathcal{I}(F_{(d_x,d_y)}^{IE} \mid ~S, r; \text{Enc}, \gamma)$ denotes the information from visual features. We formulated the lower bound of $I_{gain}$ as $\Omega$, which reflects the gap between visual and motif-sequence features, $ \Omega~\text{=}~\mathcal{I}(I_{x,y} \mid I,r;\text{Enc}^{{Image}}) \texttt{-}~\mathcal{I}(F_{(d_x,d_y)}^{M} \mid~S,r;\text{Enc})$ This highlights the need for image enhanced motif sequences to better capture motif relationships. Leveraging visual features, $F_{(d_x,d_y)}^{IE}$) integrates spatial and structural features, outperforming $F_{(d_x,d_y)}^{M}$. Please see Appendix F for more detailed proof. In the following section, We validated image fusion effectiveness through ablation and interpretability experiments.

\begin{sidewaystable*}[htbp]
    \centering
    \small
    \centering

    \begin{tabular}{lcccccccc}
        \toprule
        \multirow{2}{*}{\textbf{Methods}} & \multicolumn{4}{c}{\textbf{Deng's dataset}} & \multicolumn{4}{c}{\textbf{Ryu's dataset}} \\
        \cmidrule{2-9} 
        & \textbf{Acc.} & \textbf{Macro-F1} & \textbf{Macro-Rec}. & \textbf{Macro-Pre.} & \textbf{Acc.} &\textbf{ Macro-F1 }& \textbf{Macro-Rec.} & \textbf{Macro-Pre.} \\

        \midrule
        DeepDDI & 0.7807 & 0.6055 & 0.5839 & 0.6611 & 0.9323 & 0.8643 & 0.8512 & 0.8928 \\
        SSI-DDI & 0.7866 & 0.4216 & 0.3896 & 0.5139 & 0.9008 & 0.6663 & 0.6287 & 0.7507 \\
        MUFFIN & 0.8269 & 0.5245 & 0.4844 & 0.6204 & 0.9510 & 0.8566 & 0.8339 & 0.8980 \\
        GoGNN & 0.8766 & 0.6938 & 0.6841 & 0.7316 & 0.9424 & 0.8589 & 0.8451 & 0.8949 \\
        MRCGNN & 0.8979 & 0.7791 & 0.7688 & 0.8101 & 0.9567 & 0.8894 & 0.8727 & 0.9221 \\
        DSN-DDI & 0.8402 & 0.6319 & 0.6042 & 0.6916 & 0.9458 & 0.8479 & 0.8242 & 0.8945 \\
        ImageMol & 0.8875 & 0.7783 & 0.7613 & 0.8272 & 0.9174 & 0.8757 & 0.8662 & 0.8993 \\
        CGIP & 0.8757 & 0.7633 & 0.7641 & 0.8172 & 0.9335 & 0.8572 & 0.8765 & 0.8847 \\
        \midrule
        ImageDDI(w/ 2D) &\textbf{ 0.9186} & \textbf{0.8483} & \textbf{0.8413} & \textbf{0.8776} & \underline{0.9654} &\textbf{ 0.9403} & \textbf{0.9324} & \textbf{0.9605} \\
        ImageDDI(w/ 3D) & \underline{0.9014} & \underline{0.8229} & \underline{0.8230} & \underline{0.8470} & \textbf{0.9656} & \underline{0.9325} & \underline{0.9289} & \underline{0.9455} \\
        ImageDDI(w/o images) & 0.8597 & 0.7203 & 0.7278 & 0.7555 & 0.9327 & 0.8829 & 0.8858 & 0.9125 \\
        ImageDDI(w/ graph pos)& 0.8326 & 0.7105 & 0.7223 & 0.7456 & 0.9410 & 0.8910 & 0.8765 & 0.9049 \\
        \bottomrule
    \end{tabular}
    \caption{Results of ImageDDI and baselines for DDI event prediction on two datasets. In this table, the \textbf{bold} denotes the best result
    , and the \underline{underline} is the second best result.}
    \label{tab: performance}
\end{sidewaystable*}

\section{Experiments}
This section presents the experimental settings, model comparison with baselines, performance on new drugs, hyperparameter sensitivity analysis, and visual case analysis. More experimental details and results are provided in the supplementary materials.

\subsection{Experimental Settings}
\noindent\textbf{Datasets.} We evaluated ImageDDI on three datasets for common and inductive DDI prediction. In common scenarios, we used \textbf{\textit{Deng’s dataset}}~\cite{deng2020multimodal} (37,159 DDIs, 567 drugs, 65 events) and \textbf{\textit{Ryu’s dataset}}~\cite{ryu2018deep} (191,075 DDIs, 1,689 drugs, 86 events). For the inductive scenario, we tested on the DrugBank dataset \cite{assempour20185} (191,808 DDIs, 1,706 drugs, 86 categories).

\noindent\textbf{Baselines}. In common DDI prediction scenarios, we compare ImageDDI with several representative DDI event prediction baselines, which can be classified as follows:
\begin{itemize}
\item \textbf{DeepDDI} \cite{ryu2018deep} utilizes structural similarity profiles and DNN to predict DDI types using drug names and structural information.
\item  \textbf{SSI-DDI} \cite{nyamabo2021ssi} utilizes a Graph Attention Network (GAT) on molecular graphs, combining multi-layer GAT embeddings with a co-attention mechanism to predict drug pair interactions.
\item \textbf{MUFFIN} \cite{chen2021muffin} is a multi-scale feature fusion model that integrates drug structure, knowledge graphs, and biological information to predict DDIs.
\item \textbf{GoGNN} \cite{wang2020gognn} uses GNNs on molecular and interaction graphs, leveraging self-attention and graph pooling to predict DDIs and CCIs.
\item \textbf{MRCGNN} \cite{xiong2023multi} is a leading DDI event prediction model that uses drug molecular and DDI event graph features, refined by contrastive learning.
\item \textbf{DSN-DDI} \cite{li2023dsn} is a novel DDI prediction framework that uses dual-view drug representation learning, alternating between local (single drug) and global (drug pair) modules to effectively predict drug interactions.
\item \textbf{ImageMol} \cite{zeng2022accurate} is an unsupervised deep learning framework pretraining on 10 million molecular images to extract chemical structures.
\item \textbf{CGIP} \cite{xiang2023chemical} is a contrastive learning framework combining graph and image to learn molecular representations from large-scale unlabeled data.
\end{itemize}
In the inductive DDI prediction scenario, we included image-based models (ImageMol, CGIP), DF-based models (CSMDDI), GF-based models (HIN-DDI), and GNN-based models (KG-DDI, DeepLGF).
\begin{itemize}
\item \textbf{CSMDDI} \cite{liu2022predict} utilizes a RESCAL-based method to embed drugs and DDIs, maps drug attributes to embeddings with partial least squares regression, and employs a random forest classifier for DDI prediction.
\item \textbf{HIN-DDI} \cite{tanvir2021predicting} models drug-biomolecular relationships via a Heterogeneous Information Network and meta-path analysis for DDI prediction.
\item \textbf{KG-DDI} \cite{lin2020kgnn} integrates diverse biomedical data via a Knowledge Graph to model complex drug relationships for DDI prediction.
\item \textbf{DeepLGF} \cite{ren2022biomedical} encodes drug structures from SMILES and uses a Biomedical Knowledge Graph to extract local-global information for DDI prediction.
\end{itemize}
\textbf{Implementation Details.} For evaluation, we split each dataset into training, validation, and test sets (7:1:2 ratio), ensuring all interaction types are represented. The task is multi-class classification, evaluated using Accuracy, Macro-F1, Macro-Recall, and Macro-Precision. Hyperparameters: learning rate = 1e-4, weight decay = 1e-6, sequence length = 16. The best model, based on Macro-F1 from the validation set, was trained for up to 100 epochs and tested. Each model ran five times with different splits, and average metrics were reported.  More details are in Appendix A. Code is available at https://github.com/1hyq/ImageDDI.
\subsection{Comparison with Baselines}

Table~\ref{tab: performance} shows the overall performance of ImageDDI. Data for DeepDDI, SSI-DDI, MUFFIN, GoGNN, and MRCGNN are from MRCGNN~\cite{xiong2023multi}, while ImageMol and CGIP results are reproduced. In the common DDI prediction scenario, we compare baseline models across two datasets. The key observations are as follows:

(1) ImageDDI (w/ 2D) improves accuracy by \textbf{16.78}\%, and ImageDDI (w/ 3D) by \textbf{14.59}\% compared to SSI-DDI, demonstrating that visual molecular representations enhance DDI prediction.
(2) Our model outperforms the MRCGNN model across all metrics, highlighting that image-enhanced motif sequences better capture global and local molecular features. ImageDDI (w/ 2D) outperforms ImageDDI (w/ 3D), likely due to the clearer representations of 2D images compared to dynamic 3D structures.
(3) While image-based models perform well, they focus on single-molecule representations, limiting their ability to capture complex drug interactions. Our model better captures these relationships, improving Macro-F1 by \textbf{10.39}\% on the Deng's dataset and \textbf{6.82}\% on the Ryu's dataset.

\begin{table}[ht]
\centering
\resizebox{0.8\columnwidth}{!}{ 
\begin{tabular}{ccccc}
\toprule
\multirow{2}{*}{\textbf{Methods}} & \multicolumn{2}{c}{\textbf{S1 Partition }} & \multicolumn{2}{c}{\textbf{S2 Partition }} \\
\cmidrule(lr){2-3} \cmidrule(lr){4-5}
 & \textbf{Macro-F1} & \textbf{Acc.} & \textbf{Macro-F1} & \textbf{Acc.}  \\
\midrule
CSMDDI & 45.5 & 62.6  & 19.8 & 37.3  \\
HIN-DDI & 37.3 & 58.9  & 8.8 & 27.6  \\
KG-DDI & 26.1 & 46.7  & 1.1 & 32.2  \\
DeepLGF & 39.7 & 60.7  & 4.8 & 31.9 \\
ImageMol & 49.5 & 59.5  & 18.9 & 36.5\\
CGIP & 48.9 & 58.5 & 23.6 & 40.8   \\
\textbf{ImageDDI(w/ 2D)} & \textbf{54.9} & \underline{66.9}  & \underline{24.4} & \textbf{43.5} \\
\textbf{ImageDDI(w/ 3D)} & \underline{53.7} & \textbf{67.3}  & \textbf{24.6} & \underline{42.9} \\
\bottomrule
\end{tabular}
\centering
}
\caption{Performance evaluation of ImageDDI compared to other baselines in the inductive setting on the DrugBank dataset. The \textbf{bold} denotes the best result and the \underline{underline} denotes the second-best result.}
\label{tab: inductive}
\end{table}
\subsection{Inductive Setting}
With the development of new drugs, especially for rare or severe diseases, these drugs often contain new substances with unknown risks and haven't been extensively regulated. Identifying (DDIs) for these drugs is crucial. To evaluate the model’s performance on unseen drugs (inductive), we partition the dataset: $d_{\text{new}}$ for new drugs, $ d_{\text{old}}$ for seen drugs, with $d_{\text{new}} \cup d_{\text{old}} = D $ and $d_{\text{new}} \cap d_{\text{old}} = \varnothing $.
The dataset split consists of:
\( \text{\textbf{S1}} = \{(d_x, d_y, r) \mid d_x \in d_{\text{new}} \land d_y \in d_{\text{new}}\} \), which predicts the interaction type between emerging and existing drugs; \( \text{\textbf{S2}} = \{(d_x, d_y, r) \mid d_x \in d_{\text{new}} \land d_y \in d_{\text{new}}\} \), which predicts the interaction type between two emerging drugs. Refer to Appendix B for details. Table~\ref{tab: inductive} compares the Macro-F1 scores and Accuracy of different methods under S1 and S2 settings, highlighting the significant advantages of ImageDDI(including both 2D and 3D images).

In the \textbf{S1} setting, ImageDDI outperforms all models, with Macro-F1 and Accuracy exceeding the best baselines (e.g., CSMDDI and ImageMol) by \textbf{9.4}\% and \textbf{7.4}\%, respectively. ImageDDI (w/ 3D) also performs well, showing its ability to leverage both global visual and motif features for better DDI prediction. In the \textbf{S2} inductive setting, ImageDDI (w/ 2D) surpasses the top baseline, CGIP, by 0.83\% and 2.75\% in Macro-F1 and accuracy. This demonstrates ImageDDI’s adaptability and robustness in predicting unseen drug interactions. Overall, ImageDDI consistently excels, proving the value of incorporating image-enhanced motif sequence.

\subsection{Ablation Study} \label{5.4}
To study the effectiveness of each component of the ImageDDI model, we considered different variants:
\begin{itemize} 
\item \textit{ImageDDI without images} (w/o images): Models motif sequences without visual information. 
\item \textit{ImageDDI with graph positional info} (w/ graph pos): Treats motifs as nodes, connects drugs via common substructures, and uses a position matrix for richer contextual representation. 
\item \textit{ImageDDI with 2D image} (w/ 2D): Uses 2D images to capture overall visual representation, enhancing sequence expressiveness. \item \textit{ImageDDI with 3D image} (w/ 3D): Uses 3D images to capture overall visual representation and dynamic changes, further enhancing sequence expressiveness. \end{itemize}

\begin{figure}[h]
    \centering
    \includegraphics[width=0.99\linewidth]{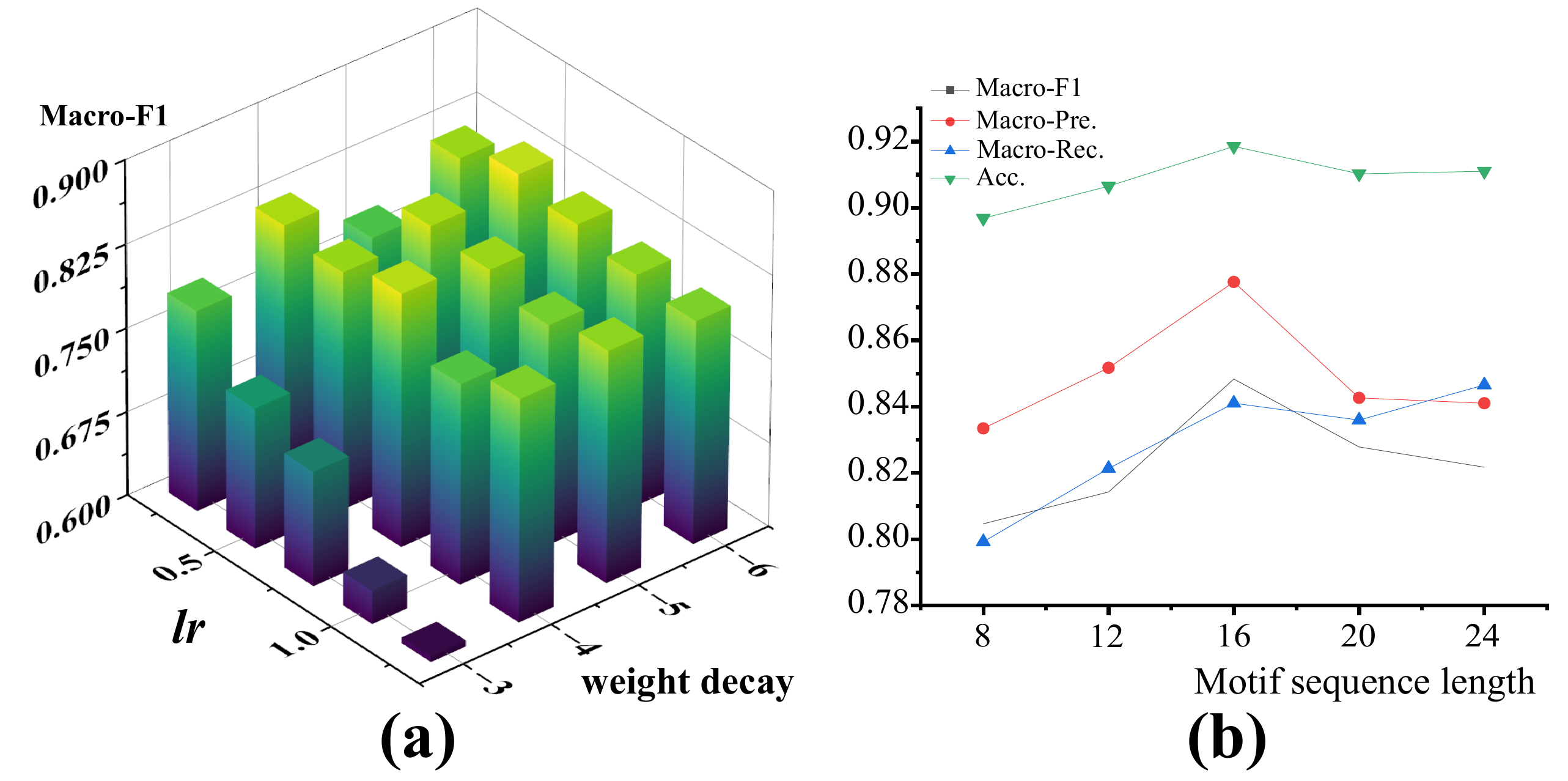}
    \caption{Hyper-parameter sensitivity analysis}
    \label{fig:hp}
\end{figure}
\noindent Experimental results (Table~\ref{tab: performance}) show that incorporating image information significantly improves performance. \textbf{ImageDDI (w/o images)} underperforms due to missing global visual information. \textbf{ImageDDI (w/ graph pos)} improves context understanding but lacks visual detail. In contrast, \textbf{ImageDDI (w/ 2D)} and \textbf{ImageDDI (w/ 3D)} capture both positions and visual details, boosting DDI prediction. However, ImageDDI (w/ 2D) outperforms ImageDDI (w/ 3D) as 2D images provide clearer visuals, while dynamic 3D frames can obscure details. Thus, ImageDDI (w/ 2D) strikes a better balance between clarity and representation.

\subsection{Hyper-parameter Sensitivity Analysis}
In this section, we conduct a hyper-parameter sensitivity analysis on ImageDDI (w/ 2D) to examine the impact of the learning rate ( \textit{lr} ), weight decay, and drug motif sequence length (\(L\)) on model performance.

\noindent\textbf{The Impact of The Combination of \textit{lr} and weight decay.} We experimented with various combinations of \textit{lr} \(\{2.5e{-4}, 5e{-4}, 7.5e{-4}, 1e{-3}, 1.25e{-3}\}\) and weight decay \(\{1e{-3}, 1e{-4}, 1e{-5}, 1e{-6}\}\). Figure~\ref{fig:hp}(a) shows the results (detailed analysis in Appendix C). The best performance was achieved with \textit{lr} = 1e{-3} and weight decay = \(1e{-6}\), showing the importance of balancing \text{lr} and weight decay for optimization and regularization, leading to higher accuracy. This analysis highlights optimal settings to prevent overfitting or underfitting and improve performance.
\subsection{Visual explanations for ImageDDI} \label{5.6}

\noindent\textbf{Effect of Drug Motif Sequence Length \( L \).} We tested drug motif sequence lengths \( L \) of 8, 12, 16, 20, and 24. As shown in Figure~\ref{fig:hp} (b), performance improves up to \( L \) =16, then declines. This suggests that an optimal sequence length captures enough information, while longer sequences add noise and shorter ones miss key details, both reducing performance.

\begin{figure}
    \centering
    \includegraphics[width=1\linewidth]{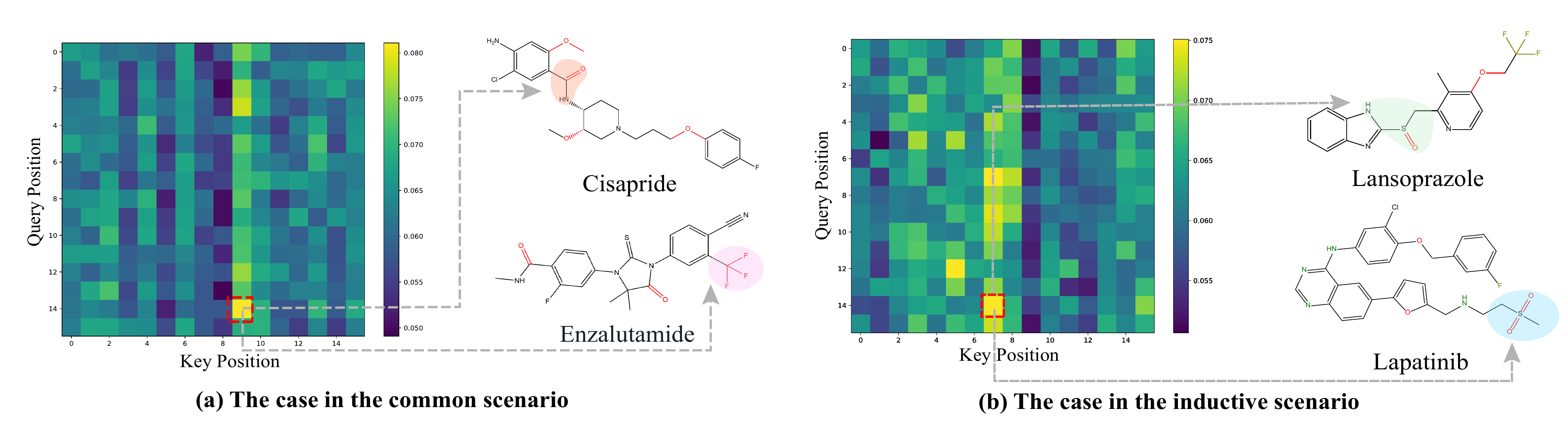}
    \caption{Motif attention weight heatmap}
    \label{fig:heatmapl}
\end{figure}
\noindent\textbf{2D Visual Explanations.} To assess the model's ability to identify key motifs in DDIs, we used heatmaps to highlight motifs with higher attention weights and compared them to existing literature. Figure~\ref{fig:heatmapl} shows two DDI cases. In Figure~\ref{fig:heatmapl} (a), the DDI between \textit{Cisapride} and \textit{Enzalutamide}\cite{wiseman1994cisapride} is shown. The heatmap highlights the amide group in \textit{Cisapride} for 5-HT4 receptor binding, aiding \textit{GERD} and \textit{gastroparesis} treatment. The trifluoromethyl group in \textit{Enzalutamide} boosts hydrophobicity, stability, and receptor affinity, enhancing anticancer efficacy. Figure~\ref{fig:heatmapl} (b) shows a cold-start DDI case with \textit{Lansoprazole} and \textit{Lapatinib}. The heatmap highlights the thiazole group in \textit{Lansoprazole}, inhibiting $\texttt{H}^{+}/\texttt{K}^{+}$\texttt{-ATPase} to treat ulcers and \textit{GERD}, and the sulfonamide group in \textit{Lapatinib}, improving EGFR and HER2 binding for better breast cancer treatment. These results show that ImageDDI accurately identifies key motifs, even in cold-start cases, matching chemical insights in DDIs. We also used t-SNE \cite{hinton2008visualizing} to visualize drug pair representations in ImageDDI (w/ 2d), as explained in Appendix D.

\noindent\textbf{3D Visual Explanations.} 
\begin{figure}
    \centering
    \includegraphics[width=0.9\linewidth]{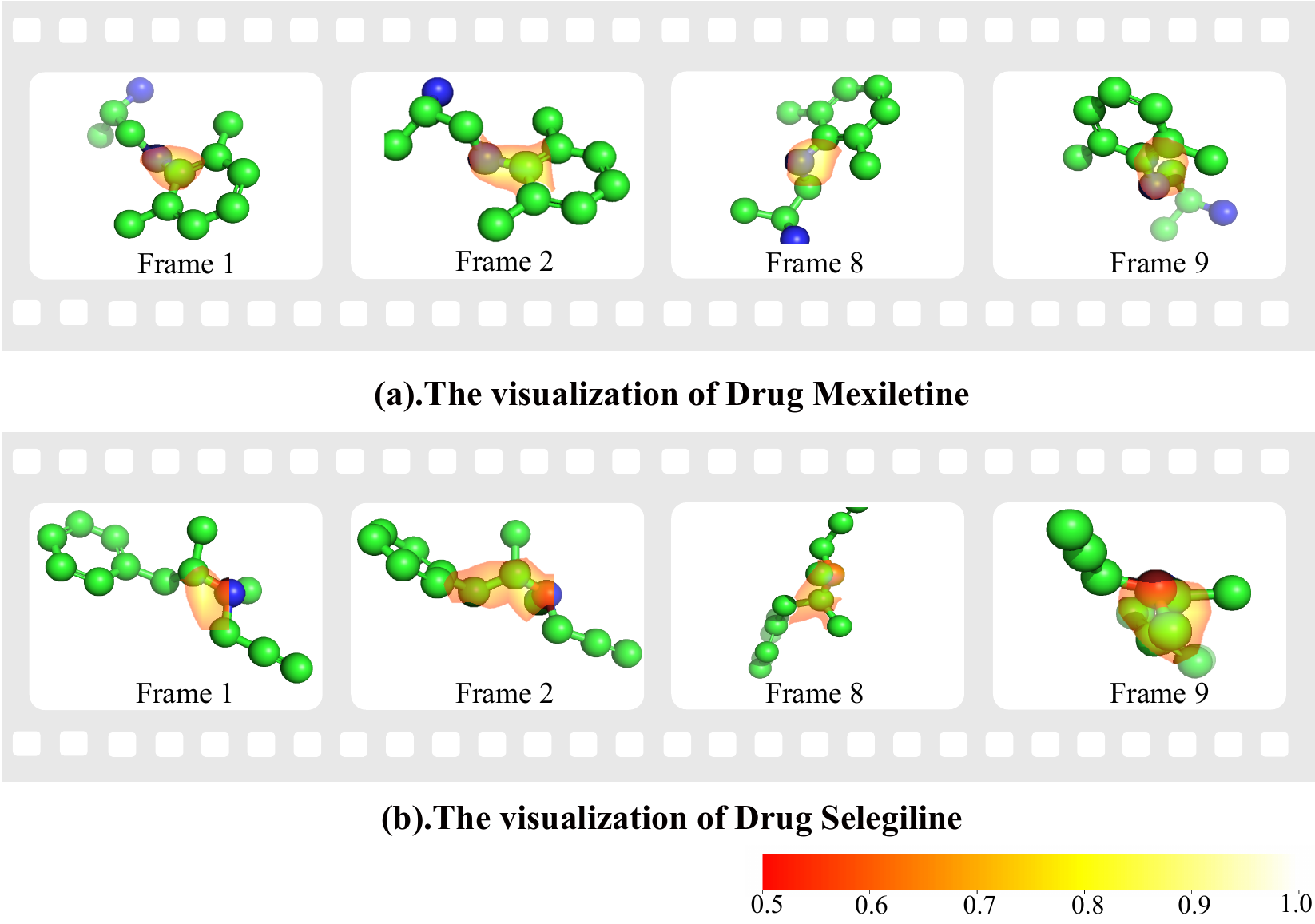}
    \caption{Case study on visualizing local attention in video frames.}
    \label{fig:enter-label}
\end{figure}
To validate the invariance of drug pairs across different video frames, we employed Grad-CAM to visualize the attention heatmaps. For the visualization process, we randomly selected four frames from ten generated by PyMol, setting attention values below 0.5 to 0. As shown in Figure~\ref{fig:enter-label}, we selected a drug pair and visualized it using the described method. The results revealed that molecules across different frames consistently focused on the same motif, demonstrating the stability and effective feature capture of our ImageDDI model (with 3D information). For the pair of drugs, our model can focus on the distinct structural features of each drug across frames, which are highlighted across all frames for each (such as the functional group \texttt{C-O-C} and amine group). These findings illustrate the model’s capability to identify critical functional motifs, demonstrating strong spatial perception and robustness.

\section{Conclusion}

This paper presents ImageDDI, a framework that enhances motif sequences with visual information to improve DDI prediction. By combining functional motifs with visual features from molecular images, ImageDDI addresses the limitations of traditional methods that focus on isolated substructure interactions. Experiments show that ImageDDI outperforms existing methods, particularly for novel drug interactions.

Future work will focus on improving generalization to unseen drugs, and developing interpretable models.

\section{Acknowledgement}
The work was supported by the National Natural Science Foundation of China (62272151, 62122025, 62425204, 62450002, 62432011 and U22A2037), the National Science and Technology Major Project (2023ZD0120902), Hunan Provincial Natural Science Foundation of China (2022JJ20016), The science and technology innovation Program of Hunan Province (2022RC1099), and the Beijing Natural Science Foundation (L248013).

\section{Appendix}

\subsection{Details about Experimental Setup} 
Our model is implemented with PyTorch 1.6.0 \cite{paszke2019pytorch}. All code was developed on an Ubuntu server equipped with 1 GPU (NVIDIA GeForce 4090), CUDA 12.2, and 12 vCPUs (Intel(R) Xeon(R) Platinum 8352V CPU @ 2.10GHz). Table 1 demonstrates all the hyper-parameters of ImageDDI.
\begin{table*}[h!]
\centering
\begin{tabular}{lll}
\toprule
\textbf{Hyper-parameter}    & \textbf{Description}                                         & \textbf{Value} \\ \midrule
$T$                         &the number of multi-head attention layers     & 6                    \\
$K$                         & the number of heads of multi-head attention                          & 8             \\

node\_hidden        & the node hidden size for the ImageDDI                    & 512            \\

$D$                         & 
the number of DDI classification types                                         & 65/86           \\
$lr$                         & the number of learning rate                                       & 0.001             \\
vcab\_size             & the number of motif types                           & 3951         \\
L                & the length of drug substructure sequence                  & 16           \\
weight\_decay               & weight decay for Adam optimizer                                      & 0.000001      \\
epoch                       & the number of training epochs                                        & 100           \\
batch\_size                 & the input batch size                                                 & 128           \\ \bottomrule
\end{tabular}
\caption{Hyper-parameters of ImageDDI}
\end{table*}

\subsection{Details of the Inductive Setting}
To evaluate our method's performance on unseen new drugs (i.e., the cold-start problem), we partitioned the dataset based on drugs rather than drug-drug interactions (DDIs). Let \( G_{\text{new}} \) represent the set of new drugs (i.e., drugs not seen during training), and \( G_{\text{old}} \) represent the seen drugs, such that \( G_{\text{new}} \cup G_{\text{old}} = G \) and \( G_{\text{new}} \cap G_{\text{old}} = \emptyset \). Based on this, the new dataset split consists of:

\begin{itemize}
    \item \textbf{Training Set \( D_{\text{train}} \)}: Includes those drug pairs used for training (i.e., seen drugs), defined as:
    \[
    D_{\text{train}} = \{(G_x, G_y, r) \in D \mid G_x \in G_{\text{old}} \land G_y \in G_{\text{old}}\}
    \]

        \item \textbf{Test Set $D_{S1}$}: Includes those pairs where at least one drug is new, defined as:
\[
\small
D_{S1} = \{(G_x, G_y, r) \in D \mid (G_x \in G_{old} \land G_y \in G_{new})
\]
\normalsize

    \item \textbf{Test Set \( D_{S2} \)}: Includes all new drug pairs (i.e., drugs not seen during training), defined as:
    \[
    D_{S1} = \{(G_x, G_y, r) \in D \mid G_x \in G_{\text{new}} \land G_y \in G_{\text{new}}\}
    \]

\end{itemize}

Specifically, \( D_{\text{train}} \) contains interactions between drugs that are all known (i.e., seen during training). \( D_{S1} \) contains entirely new drug pairs, which the model has never encountered during training. For \( D_{S2} \), it includes pairs where some drugs are new, and some are old.

The model is first trained on \( D_{\text{train}} \), then tested on both \( D_{S1} \) and \( D_{S2} \). This setup allows us to evaluate the model's ability to predict drug-drug interactions involving unseen drugs, particularly in cold-start scenarios.The inductive setting is evaluated on the DrugBank dataset, with the specific dataset partitioning details shown in Table 2.

\subsection{Detailed results of the Hyper-parameter experimental setup} 
During model training, learning rate (\(lr\)) and weight decay are critical hyperparameters. The learning rate controls the step size for updating model parameters, while weight decay acts as a regularization technique to prevent overfitting. We evaluated the model using Accuracy, Macro-Recall, Macro-precision, and Macro-F1; the final results for Macro-Recall, Macro-Precision, and Accuracy are presented in Figure 1.

To identify the best combination of learning rate and weight decay, we conducted extensive experiments. We tested various combinations of \(lr\) and weight decay, using learning rates of \(\{2.5e{-4}, 5e{-4}, 7.5e{-4}, 1e{-3}, 1.25e{-3}\}\) and weight decay values of \(\{1e{-3}, 1e{-4}, 1e{-5}, 1e{-6}\}\). The results showed that the model performed best with a learning rate of \(1 \times 10^{-3}\) and weight decay of \(1 \times 10^{-6}\), indicating that this combination optimally balances learning and regularization, leading to improved prediction accuracy.

The choice of these hyperparameters significantly impacts the model's convergence speed and generalization ability. A high learning rate may cause instability, while a low rate can lead to slow convergence. Similarly, too much weight decay can cause underfitting, while too little may not adequately prevent overfitting.

Overall, our analysis highlights the importance of fine-tuning these hyperparameters to achieve optimal model performance.

\begin{sidewaystable*}[htbp]
\centering
\small
\setlength{\tabcolsep}{4pt} 
\begin{tabular}{c|c|cccccc|cccccc}
\toprule
\multirow{2}{*}{\textbf{Data}} & \multirow{2}{*}{\textbf{seed}} & \multicolumn{6}{c|}{S1} & \multicolumn{6}{c}{S2} \\
 &  & {\text{D\_train}} & {\text{D\_valid}} & {\text{D\_test}} & {\text{S\_train}} & {\text{S\_valid}} & {\text{S\_test}} &{\text{D\_train}} & {\text{D\_valid}} & {\text{D\_test}} & {\text{S\_train}} &{\text{S\_valid}} & {\text{S\_test}} \\
\midrule
\multirow{5}{*}{DrugBank} 
 & 1 & 1461 & 1375 & 1495 & 137864 & 17591 & 32322 & 1461 & 71 & 152 & 137864 & 536 & 1901 \\
 & 12 & 1465 & 1325 & 1462 & 140085 & 17403 & 30731 & 1465 & 69 & 149 & 140085 & 522 & 1609 \\
 & 123 & 1466 & 1357 & 1455 & 140353 & 14933 & 32845 & 1466 & 73 & 142 & 140353 & 396 & 1964 \\
 & 1234 & 1463 & 1298 & 1445 & 139141 & 15635 & 33254 & 1463 & 72 & 143 & 139131 & 434 & 1956 \\
 & 12345 & 1461 & 1295 & 1500 & 133394 & 17784 & 35803 & 1461 & 67 & 150 & 133396 & 546 & 2355 \\
\bottomrule
\end{tabular}
\caption{The statistics of DrugBank for inductive settings, where D represents the drug set and S represents the DDI set.}
\end{sidewaystable*}

\begin{figure*}
    \centering
    \includegraphics[width=1.0\linewidth]{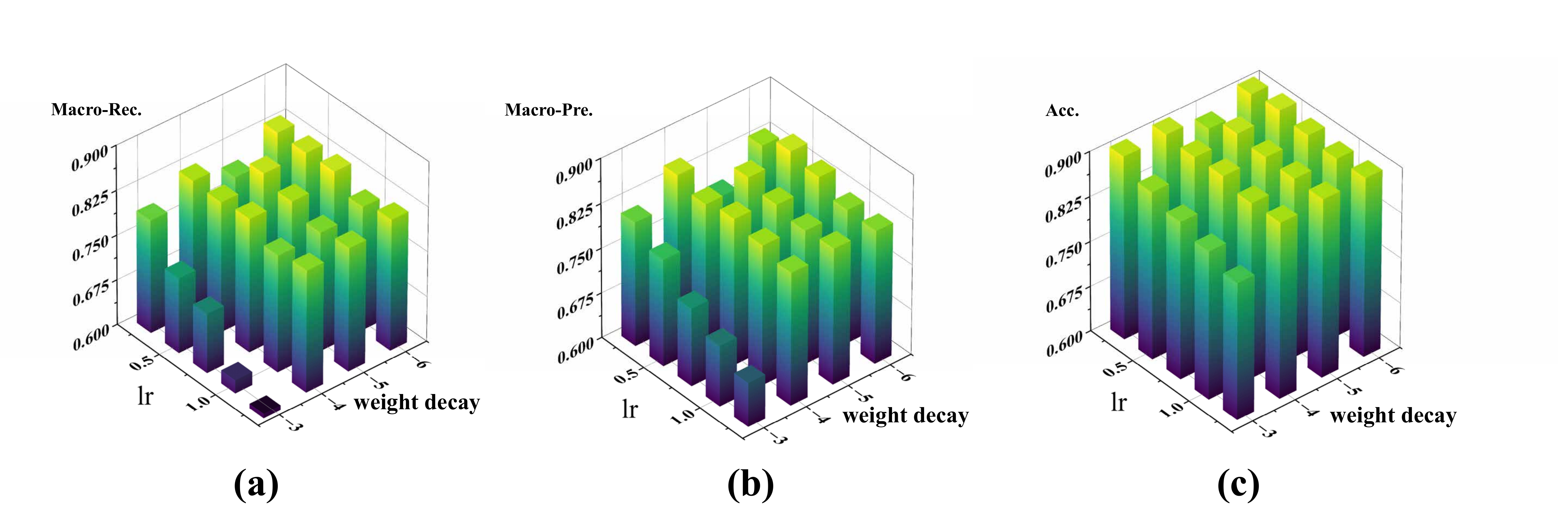}
    \caption{Results of Hyper-parameters}
    \label{fig:enter-label}
\end{figure*}

\subsection{Visualization of Deng’s dataset using the t-SNE} 
\begin{figure}
    \centering
    \includegraphics[width=0.9\linewidth]{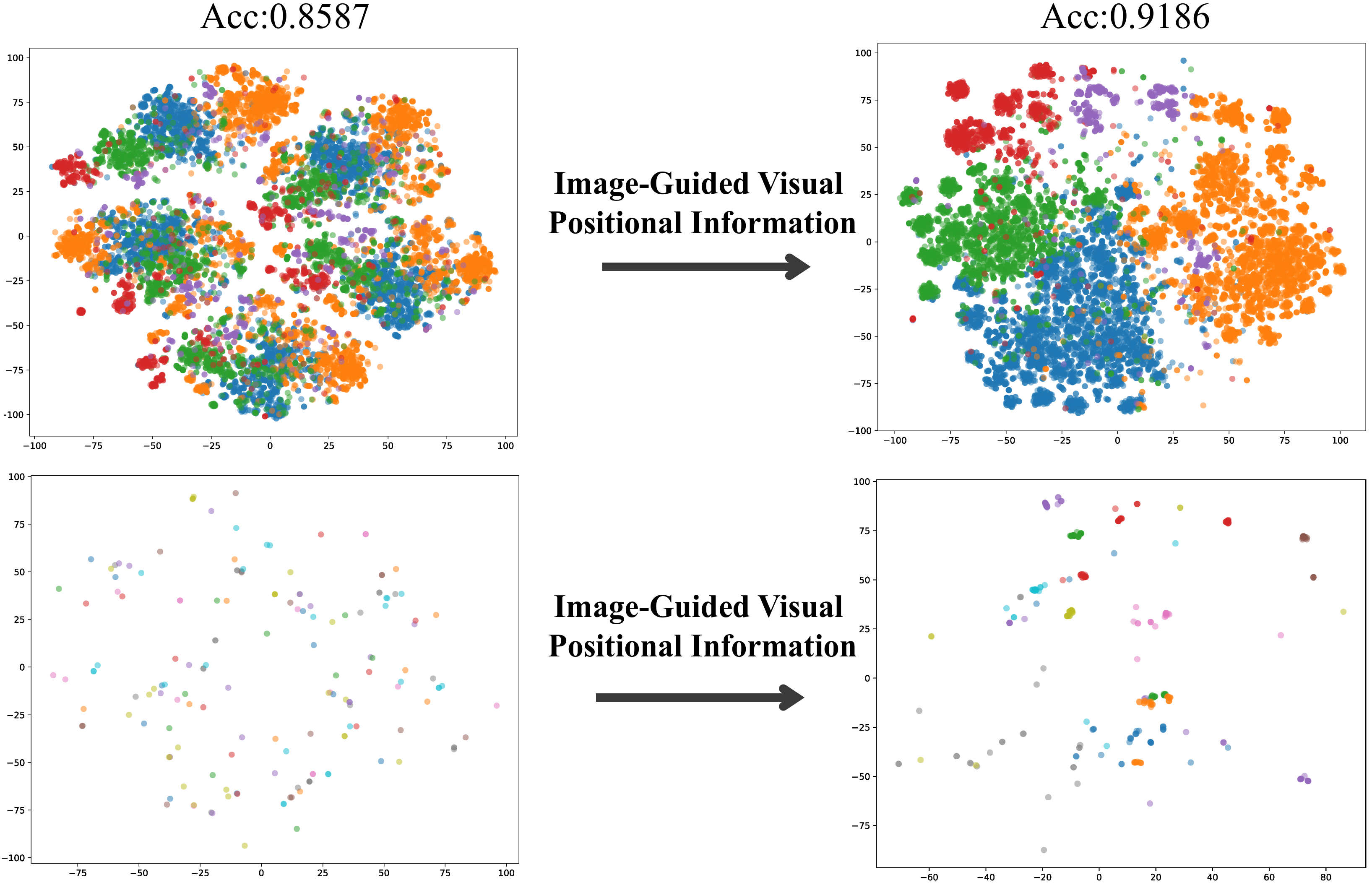}
    \caption{Visualization of Deng’s dataset using the t-SNE. 
    }
    \label{fig:t-SNE}
\end{figure}
To evaluate the impact of 2D-visual location information, we used t-SNE \cite{hinton2008visualizing} to visualize the representations of drug pairs  in ImageDDI (w/ 2D), selecting 20 low-frequency and 5 high-frequency DDI events. Figure~\ref{fig:t-SNE}
shows the clustering results, with high-frequency events in the top and low-frequency events in the bottom. Drug pairs with visual information cluster more tightly than those without one, indicating that ImageDDI(w/ 2D) effectively integrates sequence and visual features for high-quality representations. Additionally, the more compact clustering in low-frequency events suggests that ImageDDI(w/ 2D) is particularly effective at predicting rare DDI events.

\subsection{The overall process of ImageDDI}
In order to clearly describe the proposed ImageDDI framework, we show the overal process in Algorithm 1.
\begin{algorithm}[H]
\caption{Training Pipeline of ImageDDI}
\KwIn{\textbf{Require:} Training set $\mathcal{D}_{\text{train}}$, validation set $\mathcal{D}_{\text{val}}$, testing set $\mathcal{D}_{\text{test}}$, motif extraction, image encoder $\text{Enc}_I$, Transformer with adaptive feature fusion.} 

1: Construct the motif vocabulary $\mathcal{V}_\text{motif}$ from the drug set D through motif extraction.\\
2:\quad \textbf{repeat}: \\
3:\quad\quad\textbf{for} $(d_x, d_y, r) \in \mathcal{D}_{\text{train}}$ (in mini-batch) \textbf{do}: \\
4:\quad \quad \quad Extract motif sequences $S_{(d_x, d_y)}$ ;\\
5:\quad \quad \quad Extract image features$I_{(x, y)} \gets \text{Enc}_I(d_x, d_y)$ ;\\
6:\quad \quad \quad Fuse $S_{(d_x, d_y)}$ and $I_{(x, y)}$ using Transformer:\\
7:\quad \quad \quad $Z^{\prime} \gets \text{Transformer}(S_{(d_x, d_y)}, I_{(x, y)}) ;$
\\
8:\quad \quad \quad Predict $r$ using MLP with $Z^{\prime}$ ;\\
9:\quad \quad \quad Compute loss $L$ ; \\
10:\quad \quad \textbf{end for} \\
11:\quad Update $\text{Enc}_I$, Transformer;\\
12:\quad \textbf{until} performance on $\mathcal{D}_{\text{val}}$ stabilizes; \\
13:\quad Evaluate the model on $\mathcal{D}_{\text{test}}$; \\
14:\quad \textbf{return} Trained ImageDDI model.
\end{algorithm}

\subsection{Proof about the lower bound of the information increment \( \mathcal{I}_\text{gain}\)}
We provide a detailed theoretical proof for the lower bound \(\Omega\) of the information increment \(\mathcal{I}_{gain}\). The information increment is defined as the increase in useful information of one feature compared to another. Let: \(
\mathcal{I}_{gain} = \mathcal{I}^{IE} - \mathcal{I}^M,\) which represents the knowledge gain after incorporating image guidance. The image-enhanced feature \(F^{IE}_{(dx,dy)}\) is related to the motif sequence feature \(F^{M}_{(dx,dy)}\). The information contained in \(F^{IE}_{(dx,dy)}\) can be formulated as: \( \mathcal{I}^{IE} = \mathcal{I}(F^{IE}_{(dx,dy)} \mid \mathcal{S}, r; Enc, \gamma),\)
where \(\mathcal{S}\) represents the motif sequence, and \(r\) is the corresponding label. Similarly, the information contained solely in the motif sequence can be expressed as: \(
\mathcal{I}^M = \mathcal{I}(F^{M}_{(dx,dy)} \mid \mathcal{S}, r; Enc). \) Assuming the image feature \(F^{image}\) and the motif feature \(F^{M}_{(dx,dy)}\) are integrated using a transform \cite{vaswani2017attention} model, we have: \(
F^{IE}_{(dx,dy)} = Transform(F^{image}, F^{M}_{(dx,dy)}),\) where \(Transform\) represents a mapping function that captures interactions between image features and substructure features. Under this framework, \(F^{IE}_{(dx,dy)}\) combines contextual information from both image features and substructure sequences. The information increment \(\mathcal{I}_{gain}\) can then be written as:
\begin{equation}
   \mathcal{I}_{diff} = \mathcal{I}(F^{IE}_{(dx,dy)} \mid \mathcal{S}, r; Enc, \gamma) - \mathcal{I}(F^{M}_{(dx,dy)} \mid \mathcal{S}, r; Enc). \tag{1} 
\end{equation}

Due to the integration of image features through the transform model, \(F^{IE}_{(dx,dy)}\) captures additional information that enhances the representation. Therefore:
\begin{equation}
\mathcal{I}(F^{IE}_{(dx,dy)} \mid \mathcal{S}, r; Enc, \gamma) \geq \mathcal{I}(F^{M}_{(dx,dy)} \mid \mathcal{S}, r; Enc),
\end{equation}

which implies \(\mathcal{I}_{gain} \geq 0\). Furthermore, the lower bound of \(\mathcal{I}_{gain}\) can be expressed as:
\begin{equation}
\mathcal{I}_{gain} \geq \Omega = \mathcal{I}(F^{image} \mid \mathcal{S}, r; Enc^{image}) - \mathcal{I}(F^{M}_{(dx,dy)} \mid \mathcal{S}, r; Enc). \tag{2}
\end{equation}

Here, \(\mathcal{I}(F^{image} \mid \mathcal{S}, r; Enc^{image})\) represents the contextual information derived directly from image features, and \(\Omega\) reflects the knowledge increment between image features and substructure sequence features.

This analysis demonstrates that integrating image features and substructure features through a transform model can significantly enhance the representation of substructure sequences. Clearly, when the transform model effectively captures the associations between image and substructure features, \(\mathcal{I}_{gain} \geq 0\) holds true.

\subsection{Details of the BRICS decomposition of molecular motifs}

The drug set \( D \) contains multiple drug molecules, each of which is processed by the \texttt{brics.decomp()} function to extract its corresponding motif set \( S_{d_i} \). \texttt{brics.decomp()} is a key function in the BRICS method. It analyzes the cleavable chemical bonds in the molecule and, following specific chemical rules, breaks the drug molecule into smaller, functionally significant fragments known as BRICS motifs. These motifs often play a crucial role in the chemical activity or biological properties of the molecule, making them highly valuable in drug design and optimization. Each motif \( m_i \) is assigned a unique identifier and added to the motif vocabulary \( V_{\text{motif}} \), ensuring the uniqueness and consistency of the data. This vocabulary records all the motifs found in the drug molecules, providing a foundation for further drug interaction studies and helping to better understand the relationship between molecular structure and function.


 \bibliography{ref}

\begin{thebibliography}{10}

\bibitem{vilar2014similarity}
Santiago Vilar, Eugenio Uriarte, Lourdes Santana, Tal Lorberbaum, George Hripcsak, Carol Friedman, and Nicholas~P Tatonetti.
\newblock Similarity-based modeling in large-scale prediction of drug-drug interactions.
\newblock {\em Nature Protocols}, 9(9):2147--2163, 2014.

\bibitem{bansal2014community}
Mukesh Bansal, Jichen Yang, Charles Karan, Michael~P Menden, James~C Costello, Hao Tang, Guanghua Xiao, Yajuan Li, Jeffrey Allen, Rui Zhong, et~al.
\newblock A community computational challenge to predict the activity of pairs of compounds.
\newblock {\em Nature Biotechnology}, 32(12):1213--1222, 2014.

\bibitem{palmer2017combination}
Adam~C Palmer and Peter~K Sorger.
\newblock Combination cancer therapy can confer benefit via patient-to-patient variability without drug additivity or synergy.
\newblock {\em Cell}, 171(7):1678--1691, 2017.

\bibitem{giacomini2007good}
Kathleen~M Giacomini, Ronald~M Krauss, Dan~M Roden, Michel Eichelbaum, Michael~R Hayden, and Yusuke Nakamura.
\newblock When good drugs go bad.
\newblock {\em Nature}, 446(7139):975--977, 2007.

\bibitem{jin2017multitask}
Bo~Jin, Haoyu Yang, Cao Xiao, Ping Zhang, Xiaopeng Wei, and Fei Wang.
\newblock Multitask dyadic prediction and its application in prediction of adverse drug-drug interaction.
\newblock In {\em Proceedings of the AAAI Conference on Artificial Intelligence}, volume~31, 2017.

\bibitem{qiu2021comprehensive}
Yang Qiu, Yang Zhang, Yifan Deng, Shichao Liu, and Wen Zhang.
\newblock A comprehensive review of computational methods for drug-drug interaction detection.
\newblock {\em IEEE/ACM Transactions on Computational Biology and Bioinformatics}, 19(4):1968--1985, 2021.

\bibitem{ryu2018deep}
Jae~Yong Ryu, Hyun~Uk Kim, and Sang~Yup Lee.
\newblock Deep learning improves prediction of drug--drug and drug--food interactions.
\newblock {\em Proceedings of the national academy of sciences}, 115(18):E4304--E4311, 2018.

\bibitem{huang2020caster}
Kexin Huang, Cao Xiao, Trong Hoang, Lucas Glass, and Jimeng Sun.
\newblock Caster: Predicting drug interactions with chemical substructure representation.
\newblock In {\em Proceedings of the AAAI Conference on Artificial Intelligence}, volume~34, pages 702--709, 2020.

\bibitem{zhang2019graph}
Si~Zhang, Hanghang Tong, Jiejun Xu, and Ross Maciejewski.
\newblock Graph convolutional networks: a comprehensive review.
\newblock {\em Computational Social Networks}, 6(1):1--23, 2019.

\bibitem{nyamabo2021ssi}
Arnold~K Nyamabo, Hui Yu, and Jian-Yu Shi.
\newblock Ssi--ddi: substructure--substructure interactions for drug--drug interaction prediction.
\newblock {\em Briefings in Bioinformatics}, 22(6):bbab133, 2021.

\bibitem{nyamabo2022drug}
Arnold~K Nyamabo, Hui Yu, Zun Liu, and Jian-Yu Shi.
\newblock Drug--drug interaction prediction with learnable size-adaptive molecular substructures.
\newblock {\em Briefings in Bioinformatics}, 23(1):bbab441, 2022.

\bibitem{yang2022learning}
Ziduo Yang, Weihe Zhong, Qiujie Lv, and Calvin Yu-Chian Chen.
\newblock Learning size-adaptive molecular substructures for explainable drug--drug interaction prediction by substructure-aware graph neural network.
\newblock {\em Chemical science}, 13(29):8693--8703, 2022.

\bibitem{zhao2023improving}
Weizhong Zhao, Xueling Yuan, Xianjun Shen, Xingpeng Jiang, Chuan Shi, Tingting He, and Xiaohua Hu.
\newblock Improving drug--drug interactions prediction with interpretability via meta-path-based information fusion.
\newblock {\em Briefings in Bioinformatics}, 24(2):bbad041, 2023.

\bibitem{zitnik2018modeling}
Marinka Zitnik, Monica Agrawal, and Jure Leskovec.
\newblock Modeling polypharmacy side effects with graph convolutional networks.
\newblock {\em Bioinformatics}, 34(13):i457--i466, 2018.

\bibitem{chen2021muffin}
Yujie Chen, Tengfei Ma, Xixi Yang, Jianmin Wang, Bosheng Song, and Xiangxiang Zeng.
\newblock Muffin: multi-scale feature fusion for drug--drug interaction prediction.
\newblock {\em Bioinformatics}, 37(17):2651--2658, 2021.

\bibitem{yu2021sumgnn}
Yue Yu, Kexin Huang, Chao Zhang, Lucas~M Glass, Jimeng Sun, and Cao Xiao.
\newblock Sumgnn: multi-typed drug interaction prediction via efficient knowledge graph summarization.
\newblock {\em Bioinformatics}, 37(18):2988--2995, 2021.

\bibitem{lin2020kgnn}
Xuan Lin, Zhe Quan, Zhi-Jie Wang, Tengfei Ma, and Xiangxiang Zeng.
\newblock Kgnn: Knowledge graph neural network for drug-drug interaction prediction.
\newblock In {\em IJCAI}, volume 380, pages 2739--2745, 2020.

\bibitem{wang2022predicting}
Fei Wang, Xiujuan Lei, Bo~Liao, and Fang-Xiang Wu.
\newblock Predicting drug--drug interactions by graph convolutional network with multi-kernel.
\newblock {\em Briefings in Bioinformatics}, 23(1):bbab511, 2022.

\bibitem{xiong2023multi}
Zhankun Xiong, Shichao Liu, Feng Huang, Ziyan Wang, Xuan Liu, Zhongfei Zhang, and Wen Zhang.
\newblock Multi-relational contrastive learning graph neural network for drug-drug interaction event prediction.
\newblock In {\em Proceedings of the AAAI Conference on Artificial Intelligence}, volume~37, pages 5339--5347, 2023.

\bibitem{wu2024mkg}
Di~Wu, Wu~Sun, Yi~He, Zhong Chen, and Xin Luo.
\newblock Mkg-fenn: A multimodal knowledge graph fused end-to-end neural network for accurate drug--drug interaction prediction.
\newblock In {\em Proceedings of the AAAI Conference on Artificial Intelligence}, volume~38, pages 10216--10224, 2024.

\bibitem{niu2024srr}
Dongjiang Niu, Lei Xu, Shourun Pan, Leiming Xia, and Zhen Li.
\newblock Srr-ddi: A drug--drug interaction prediction model with substructure refined representation learning based on self-attention mechanism.
\newblock {\em Knowledge-Based Systems}, 285:111337, 2024.

\bibitem{tang2023dsil}
Zhenchao Tang, Guanxing Chen, Hualin Yang, Weihe Zhong, and Calvin Yu-Chian Chen.
\newblock Dsil-ddi: a domain-invariant substructure interaction learning for generalizable drug--drug interaction prediction.
\newblock {\em IEEE Transactions on Neural Networks and Learning Systems}, 2023.

\bibitem{vaswani2017attention}
Ashish Vaswani, Noam Shazeer, Niki Parmar, Jakob Uszkoreit, Llion Jones, Aidan~N Gomez, {\L}ukasz Kaiser, and Illia Polosukhin.
\newblock Attention is all you need.
\newblock {\em Advances in neural information processing systems}, 30, 2017.

\bibitem{morgan1965generation}
Harry~L Morgan.
\newblock The generation of a unique machine description for chemical structures-a technique developed at chemical abstracts service.
\newblock {\em Journal of chemical documentation}, 5(2):107--113, 1965.

\bibitem{velivckovic2017graph}
Petar Veli{\v{c}}kovi{\'c}, Guillem Cucurull, Arantxa Casanova, Adriana Romero, Pietro Lio, and Yoshua Bengio.
\newblock Graph attention networks.
\newblock {\em arXiv preprint arXiv:1710.10903}, 2017.

\bibitem{guo2024dsil}
Lantu Guo, Jun Lu, Jianping An, and Kai Yang.
\newblock Dsil: An effective spectrum prediction framework against spectrum concept drift.
\newblock {\em IEEE Transactions on Cognitive Communications and Networking}, 2024.

\bibitem{zhu2025drug}
Zhiqin Zhu, Yan Ding, Guanqiu Qi, Baisen Cong, Yuanyuan Li, Litao Bai, and Xinbo Gao.
\newblock Drug--target affinity prediction using rotary encoding and information retention mechanisms.
\newblock {\em Engineering Applications of Artificial Intelligence}, 147:110239, 2025.

\bibitem{zhu2024drug}
Zhiqin Zhu, Xin Zheng, Guanqiu Qi, Yifei Gong, Yuanyuan Li, Neal Mazur, Baisen Cong, and Xinbo Gao.
\newblock Drug--target binding affinity prediction model based on multi-scale diffusion and interactive learning.
\newblock {\em Expert Systems with Applications}, 255:124647, 2024.

\bibitem{zhu2023drug}
Zhiqin Zhu, Zheng Yao, Xin Zheng, Guanqiu Qi, Yuanyuan Li, Neal Mazur, Xinbo Gao, Yifei Gong, and Baisen Cong.
\newblock Drug--target affinity prediction method based on multi-scale information interaction and graph optimization.
\newblock {\em Computers in Biology and Medicine}, 167:107621, 2023.

\bibitem{zhu2023associative}
Zhiqin Zhu, Zheng Yao, Guanqiu Qi, Neal Mazur, Pan Yang, and Baisen Cong.
\newblock Associative learning mechanism for drug-target interaction prediction.
\newblock {\em CAAI Transactions on Intelligence Technology}, 8(4):1558--1577, 2023.

\bibitem{zeng2022accurate}
Xiangxiang Zeng, Hongxin Xiang, Linhui Yu, Jianmin Wang, Kenli Li, Ruth Nussinov, and Feixiong Cheng.
\newblock Accurate prediction of molecular properties and drug targets using a self-supervised image representation learning framework.
\newblock {\em Nature Machine Intelligence}, 4(11):1004--1016, 2022.

\bibitem{xiang2023chemical}
Hongxin Xiang, Shuting Jin, Xiangrong Liu, Xiangxiang Zeng, and Li~Zeng.
\newblock Chemical structure-aware molecular image representation learning.
\newblock {\em Briefings in Bioinformatics}, 24(6):bbad404, 2023.

\bibitem{10375706}
Xiang Zhang, Hongxin Xiang, Xixi Yang, Jingxin Dong, Xiangzheng Fu, Xiangxiang Zeng, Haowen Chen, and Keqin Li.
\newblock Dual-view learning based on images and sequences for molecular property prediction.
\newblock {\em IEEE Journal of Biomedical and Health Informatics}, 28(3):1564--1574, 2024.

\bibitem{he2016deep}
Kaiming He, Xiangyu Zhang, Shaoqing Ren, and Jian Sun.
\newblock Deep residual learning for image recognition.
\newblock In {\em Proceedings of the IEEE conference on computer vision and pattern recognition}, pages 770--778, 2016.

\bibitem{deng2020multimodal}
Yifan Deng, Xinran Xu, Yang Qiu, Jingbo Xia, Wen Zhang, and Shichao Liu.
\newblock A multimodal deep learning framework for predicting drug--drug interaction events.
\newblock {\em Bioinformatics}, 36(15):4316--4322, 2020.

\bibitem{assempour20185}
N~Assempour, I~Iynkkaran, Y~Liu, A~Maciejewski, N~Gale, A~Wilson, L~Chin, R~Cummings, D~Le, A~Pon, et~al.
\newblock 5.0: a major update to the drugbank database for 2018.
\newblock {\em Nucleic Acids Res}, 46:D1074--82, 2018.

\bibitem{wang2020gognn}
Hanchen Wang, Defu Lian, Ying Zhang, Lu~Qin, and Xuemin Lin.
\newblock Gognn: Graph of graphs neural network for predicting structured entity interactions.
\newblock {\em arXiv preprint arXiv:2005.05537}, 2020.

\bibitem{li2023dsn}
Zimeng Li, Shichao Zhu, Bin Shao, Xiangxiang Zeng, Tong Wang, and Tie-Yan Liu.
\newblock Dsn-ddi: an accurate and generalized framework for drug--drug interaction prediction by dual-view representation learning.
\newblock {\em Briefings in Bioinformatics}, 24(1):bbac597, 2023.

\bibitem{liu2022predict}
Zun Liu, Xing-Nan Wang, Hui Yu, Jian-Yu Shi, and Wen-Min Dong.
\newblock Predict multi-type drug--drug interactions in cold start scenario.
\newblock {\em BMC bioinformatics}, 23(1):75, 2022.

\bibitem{tanvir2021predicting}
Farhan Tanvir, Muhammad Ifte~Khairul Islam, and Esra Akbas.
\newblock Predicting drug-drug interactions using meta-path based similarities.
\newblock In {\em 2021 IEEE Conference on Computational Intelligence in Bioinformatics and Computational Biology (CIBCB)}, pages 1--8. IEEE, 2021.

\bibitem{ren2022biomedical}
Zhong-Hao Ren, Zhu-Hong You, Chang-Qing Yu, Li-Ping Li, Yong-Jian Guan, Lu-Xiang Guo, and Jie Pan.
\newblock A biomedical knowledge graph-based method for drug--drug interactions prediction through combining local and global features with deep neural networks.
\newblock {\em Briefings in bioinformatics}, 23(5):bbac363, 2022.

\bibitem{wiseman1994cisapride}
Lynda~R Wiseman and Diana Faulds.
\newblock Cisapride: an updated review of its pharmacology and therapeutic efficacy as a prokinetic agent in gastrointestinal motility disorders.
\newblock {\em Drugs}, 47:116--152, 1994.

\bibitem{hinton2008visualizing}
G~Hinton and L~Van Der~Maaten.
\newblock Visualizing data using t-sne journal of machine learning research.
\newblock {\em Journal of Machine Learning Research}, 9:2579--2605, 2008.

\bibitem{paszke2019pytorch}
Adam Paszke, Sam Gross, Francisco Massa, Adam Lerer, James Bradbury, Gregory Chanan, Trevor Killeen, Zeming Lin, Natalia Gimelshein, Luca Antiga, et~al.
\newblock Pytorch: An imperative style, high-performance deep learning library.
\newblock {\em Advances in neural information processing systems}, 32, 2019.

\end{thebibliography}
\end{document}